\documentclass[11pt]{article}

\usepackage{float}

\usepackage{amsmath,amssymb}
\usepackage{graphicx}
\usepackage{booktabs}
\usepackage{multirow}
\usepackage{hyperref}
\usepackage{algorithm}
\usepackage{algorithmic}
\usepackage{listings}
\usepackage{xcolor}
\usepackage{geometry}
\usepackage[authoryear]{natbib}   
\title{\textbf{SAFE: Stable Alignment Finetuning with Entropy-Aware Predictive Control for RLHF}\\
\large  Combining Double Soft-Min Critics with Adaptive KL Thresholding}

\author{Dipan Maity \\
\texttt{dipanai.xyz@gmail.com}}

\date{}

\begin{document}
\maketitle


\begin{abstract}\noindent Proximal Policy
Optimization (PPO) has been positioned by recent literature as the canonical method for the RL
part of Reinforcement Learning from Human Feedback (RLHF). PPO performs well empirically but has a heuristic motivation and handles the KL-divergence constraint
used in LM-RLHF in an ad-hoc manner and suffers from reward oscillations, entropy collapse, value function drift,
and sudden policy divergence that require frequent restarts and extensive hyperparameter tuning. In this paper, we develop  a new pure on policy actor-critic RL method
for the LM-RLHF setting. We present SAFE (Stable Alignment Finetuning 
with Entropy-aware control),a novel RLHF algorithm that combines a Double Soft-Min Critic for pessimistic value estimation with a new  multi-layer stabilization framework combining  entropy-gated KL regulation, and  Proportional-Integral-Derivative (PID)-controlled adaptive 
thresholds. Unlike standard PPO's symmetric KL penalties, SAFE distinguishes high-entropy 
exploration from low-entropy mode collapse and adjusts penalties dynamically based on 
reward velocity. Experiments on a \textbf{3B parameter model} show SAFE achieves 
\textbf{+5.15\%  training-average reward} than PPO (0.725 vs 0.689), \textbf{negligible reward crashes}, 
and superior KL control than ppo . Our method adds minimal computational overhead and provides an 
interpretable, crash-resistant RLHF framework that maintains aggressive learning speed 
while ensuring stable long-horizon optimization suitable for production deployment. Code is available at \url{https://github.com/ryyzn9/SAFE}

\end{abstract}



\section{Introduction}

Reinforcement Learning from Human Feedback (RLHF) \cite{ouyang2022traininglanguagemodelsfollow,bai2022traininghelpfulharmlessassistant,christiano2023deepreinforcementlearninghuman} has become a central paradigm for aligning large language models with human preferences. Despite its empirical success, RLHF training remains highly unstable, particularly in long-horizon on-policy optimization. Practitioners routinely observe failure modes including reward oscillations, entropy collapse, value function drift, and sudden policy divergence that require frequent restarts and extensive hyperparameter tuning. 

Most existing stabilization approaches focus on single-axis control, typically through fixed or adaptive Kullback–Leibler (KL) \cite{schulman2017proximalpolicyoptimizationalgorithms} regularization that constrains policy deviation from a reference model. While effective at limiting distributional drift, KL-based control alone does not fully capture internal policy dynamics. In practice, policies may remain within acceptable KL bounds while undergoing progressive determinization, increased gradient sensitivity, or unstable value estimation, leading to degraded learning dynamics and brittle convergence behavior.

A critical but often overlooked source of instability is \emph{value overestimation} in the critic network. Standard actor-critic methods rely on a single value estimator that, due to bootstrapping and function approximation error, systematically overestimates returns in high-variance regions. When the policy encounters outlier rewards—whether from genuine improvement or spurious reward model artifacts—an overestimating critic amplifies these signals through inflated advantage estimates, driving overly aggressive policy updates that can trigger cascading instability. This positive feedback loop is particularly destructive in RLHF, where reward models themselves contain systematic biases and the policy operates over enormous discrete action spaces with delayed sparse feedback.

These challenges become particularly pronounced in long training runs, where delayed instabilities often emerge after hundreds or thousands of optimization steps. Such late-stage failures are difficult to detect using instantaneous metrics and are not adequately addressed by static regularization schedules or symmetric divergence penalties. Addressing these coupled failure modes requires coordinated intervention across multiple levels of the optimization process.

In this work, we introduce \textbf{SAFE} (\textbf{S}table \textbf{A}lignment \textbf{F}inetuning with \textbf{E}ntropy-aware control), a stabilized RLHF optimization framework that integrates complementary control mechanisms across three interacting layers: value estimation, divergence regulation, and training dynamics adaptation. Rather than relying on a single stabilizer, SAFE combines pessimistic value aggregation, entropy-conditioned divergence control, and predictive threshold adaptation into a unified training architecture.

\paragraph{Contributions.} Our contributions are:

\begin{itemize}
    \item We identify and characterize interacting instability modes in long-horizon RLHF training—including value overestimation, distributional drift, and entropy collapse—that are not adequately addressed by standard KL regularization alone.
    
    \item We propose SAFE, a multi-layer stabilization framework that integrates \emph{pessimistic double soft-min value estimation} to suppress overconfident updates, \emph{entropy-aware predictive control} that couples asymmetric divergence regulation with adaptive entropy-based thresholding, and \emph{reward-driven PID adaptation} to regulate exploration dynamics across training phases.
    
    \item We demonstrate empirically that coordinated control across value, policy, and temporal dynamics improves training stability and robustness without sacrificing reward performance, and provide theoretical analysis suggesting SAFE may offer partial resistance to reward hacking through its pessimistic critic mechanism.
\end{itemize}

Together, these results suggest that robust RLHF optimization requires explicit multi-level control of value estimation, divergence dynamics, and exploration behavior rather than static regularization alone.

\section{Background }

\subsection{RLHF Pipeline}

Reinforcement Learning from Human Feedback (RLHF) typically consists of three sequential stages \cite{ouyang2022traininglanguagemodelsfollow}:

\begin{enumerate}
\item \textbf{Supervised Fine-Tuning (SFT):} A pretrained language model is adapted to follow instructions using curated human demonstrations.
\item \textbf{Reward Model Training:} A separate model is trained to approximate human preference judgments over generated responses.
\item \textbf{Reinforcement Learning Optimization:} The policy is optimized to maximize reward model outputs while remaining close to the supervised reference policy.
\end{enumerate}

The resulting optimization objective is commonly formulated as:

\begin{equation}
\mathcal{L}(\theta)
=
-\mathbb{E}_{x \sim D,\, y \sim \pi_\theta}
\big[
r(x,y)
\big]
+
\beta \cdot
D_{\text{KL}}
\big(
\pi_\theta \;\|\; \pi_{\text{ref}}
\big),
\label{eq:rlhf_obj}
\end{equation}

where $r(x,y)$ denotes the learned reward model score, $\pi_{\text{ref}}$ is the frozen supervised policy, and $\beta$ controls the strength of divergence regularization.

This KL constraint plays a dual role: it stabilizes optimization by limiting large policy updates and preserves linguistic quality by anchoring the learned policy to the pretrained distribution.

Despite its success, PPO-based RLHF remains sensitive to entropy collapse, reward hacking, and late-stage instability, motivating more structured control mechanisms.

\subsection{Log-Probability Ratio Estimation in RLHF}

\label{sec:logratio_background}

In practical RLHF implementations, KL divergence is not computed exactly over the full vocabulary distribution. Instead, divergence is estimated using Monte Carlo sampling over tokens generated by the current policy.\cite{ouyang2022traininglanguagemodelsfollow,bai2022traininghelpfulharmlessassistant}.

Specifically, the commonly used estimator is the batch mean of per-token log-probability ratios:\cite{ouyang2022traininglanguagemodelsfollow}.

\begin{equation}
\widehat{D}_{\text{KL}}
=
\frac{1}{N}
\sum_{i=1}^{N}
\Big[
\log \pi_\theta(a_i \mid s_i)
-
\log \pi_{\text{ref}}(a_i \mid s_i)
\Big],
\quad a_i \sim \pi_\theta.
\label{eq:kl_estimate}
\end{equation}

While the true KL divergence
$D_{\text{KL}}(\pi_\theta \| \pi_{\text{ref}})$
is always non-negative by definition, this finite-sample estimator can assume negative values due to sampling variance and estimator bias.

Negative estimates occur when sampled tokens have lower probability under the current policy than under the reference model on average. This phenomenon is especially common during high-entropy exploration phases, where the policy samples broadly from low-probability tail regions of the distribution.

Importantly, such negative values do not indicate negative KL divergence, but instead reflect stochastic sampling effects. Nevertheless, we observe empirically that the sign and magnitude of this estimator correlate strongly with policy behavior:

\begin{itemize}
\item \textbf{Negative log-ratio estimates} typically coincide with high-entropy exploratory regimes.
\item \textbf{Positive log-ratio estimates} reflect increasing policy confidence and concentration on high-reward modes.
\end{itemize}

Sustained positive log-ratio estimates combined with decreasing entropy are frequently observed prior to exploitative behavior and reward hacking. This motivates using the log-ratio signal jointly with entropy to regulate policy stabilization.

\subsection{Double Critic Learning and Pessimistic Aggregation}

Value overestimation is a fundamental source of instability in actor--critic
algorithms and arises from bootstrapped temporal-difference updates combined
with function approximation error. When a single critic systematically
overestimates action values, policy updates become overly aggressive,
amplifying optimization noise and often leading to training divergence.

Double Q-learning mitigates this issue by maintaining two independently
parameterized critics and applying pessimistic aggregation when computing
Q-value targets \cite{van2010double,fujimoto2018addressingfunctionapproximation}. Instead of relying on a single
value estimate, the minimum of the two critics is used to form training
targets:
\begin{equation}
V_{\text{target}}(s) = \min(Q_1(s), Q_2(s)).
\end{equation}

This pessimistic selection reduces positive bias by lowering the probability
that both critics simultaneously overestimate the same state or action.
Although this introduces a small underestimation bias, empirical studies have
shown that this trade-off substantially improves stability and robustness
across a wide range of continuous control benchmarks \cite{fujimoto2018addressingfunctionapproximation}.

Twin-critic architectures have since become a standard component of modern
off-policy and on-policy actor--critic algorithms, particularly in settings
with high variance rewards and non-stationary policy updates. The core
principle of pessimistic aggregation provides a simple yet effective mechanism
for stabilizing value estimation in large-scale reinforcement learning systems.

\subsection{Entropy Regularization and Exploration Stability}

Entropy regularization is commonly used in reinforcement learning to encourage exploration by penalizing overly deterministic policies. In maximum-entropy frameworks such as Soft Actor-Critic \cite{haarnoja2018softactorcriticoffpolicymaximum}, entropy is treated as a first-class optimization objective rather than a secondary regularizer.\cite{haarnoja2018softactorcritic,geist2019theoryregularizedmdps,
neu2017unifiedentropy,nachum2017bridging}.

In RLHF, however, entropy regularization is typically implemented using static coefficients. Such static entropy bonuses are insufficient to prevent premature entropy collapse during early training and can destabilize late-stage convergence when over-amplified.

Language model alignment exacerbates this problem due to extremely large action spaces, delayed sequence-level rewards, and strong pretrained inductive biases. Once entropy collapses prematurely, policies rapidly overfit reward model artifacts and lose the ability to recover through exploration.

These observations motivate adaptive entropy-aware stabilization mechanisms that dynamically regulate exploration based on training state rather than relying on fixed entropy coefficients.

\section{Method:}
\label{sec:asymmetric}
\paragraph{Value-Level Stabilization via Pessimistic Aggregation.} 
At the foundation of SAFE is a \emph{double soft-min critic} architecture that directly addresses value overestimation bias. Following the principle of pessimistic value estimation \cite{van2010double}, we maintain two independently parameterized critics and aggregate their estimates using a differentiable soft-minimum operator:
\[
V_{\text{soft}}(s) = -\alpha \log \left[\frac{1}{2}\left(e^{-V_1(s)/\alpha} + e^{-V_2(s)/\alpha}\right)\right].
\]
This soft aggregation provides a smooth, pessimistic lower bound that reduces the probability of both critics simultaneously overestimating the same state. Unlike hard minimum operators that can introduce gradient discontinuities, the soft-min formulation preserves differentiability while systematically suppressing optimistic outliers. Crucially, this pessimistic bias makes the policy \emph{less responsive to reward spikes}—including those arising from reward model exploitation—thereby providing a first line of defense against both training instability and reward hacking. We further stabilize critic learning through layer normalization applied to shared representations and Polyak-averaged target networks for bootstrap stability.

\paragraph{Policy-Level Stabilization via Entropy-Aware Predictive Control.} At the policy level, standard symmetric KL penalties fail because they penalize healthy exploratory deviations equally with dangerous overconfident drift. To address this fundamental limitation, SAFE introduces an \emph{entropy-aware predictive controller} that integrates three complementary mechanisms into a unified control architecture: (1) \emph{asymmetric penalization} that selectively suppresses positive KL divergence (indicating deterministic drift toward reward model artifacts) while preserving negative deviations associated with high-entropy exploration, (2) \emph{entropy-gated scaling} that amplifies corrective pressure precisely when policy entropy drops into low-stochasticity regimes characteristic of mode collapse and exploitation, and (3) \emph{ Proportional-Integral-Derivative (PID)-driven adaptive thresholding} that dynamically adjusts divergence tolerance based on reward velocity rather than fixed schedules. By coupling directional divergence control with entropy dynamics and temporal reward trends, this controller detects exploitative behavior earlier and maintains stable regulation across varying training phases.

Importantly, our approach does not impose hard entropy targets or rigid divergence constraints. Instead, SAFE adapts its control signals continuously in response to observed training behavior, allowing smooth transitions from exploratory optimization to stable convergence without introducing additional off-policy corrections or auxiliary objectives.

\subsection{Motivation: Late-Training Entropy  Collapse in RLHF}
\label{sec:asym_kl}
In RLHF, KL divergence measures how much the policy model (the model being trained) has deviated from the reference model (the original, pre-trained model). The formula used per-token is:

\begin{equation}
D_{\mathrm{KL}}(\pi \,\|\, \pi_{\mathrm{ref}}) 
= \log \pi(x_t) - \log \pi_{\mathrm{ref}}(x_t)
\end{equation}

Averaged across tokens and batches, this gives a single scalar KL value per training step. 

A critical instability in long-horizon RLHF is \emph{late-stage policy  entropy collapse}: policies initially optimize successfully but undergo abrupt divergence characterized by rapid KL growth, entropy collapse, and catastrophic reward degradation. We observed the following failure pstandard KLdouble soft critic with standered kl experiments:

\begin{table}[ht]
\centering
\small
\begin{tabular}{ccc}
\toprule
Step & KL Estimate & Reward \\
\midrule
1400 & $-0.09$ & $0.71$ \\
1450 & $+0.12$ & $0.70$ \\
1500 & $+1.36$ & $0.59$ \\
1550 & $+2.89$ & $0.41$ \\
\bottomrule
\end{tabular}
\caption{Example collapse trajectory showing transition from exploration ($\text{KL}<0$) to exploitative divergence ($\text{KL}>0$) with 42\% reward loss.}
\label{tab:collapse}
\end{table}

This transition corresponds to a shift from exploratory behavior toward narrow exploitation of reward model artifacts, followed by irreversible policy degeneration. The problem is not negative KL itself. The problem is that the adaptive controller INTERPRETS negative KL as "everything is fine, remove constraints . To prevent this we suggest making KL always positive, but   always positive KL regularization fails to prevent this because symmetric penalties of the form
\begin{equation}
L_{\text{KL}} = \beta \cdot |\hat{D}_{KL} - D_{\text{target}}|
\end{equation}
suffer from two limitations: (1) negative KL estimates associated with healthy exploration are penalized equally to dangerous positive divergence, and (2) penalties react only to instantaneous magnitude, failing to detect accelerating trends that precede collapse.

\subsection{Asymmetric Controller Design}

To address these limitations, we introduce an asymmetric controller that selectively penalizes over-confident divergence while preserving exploration and detecting rising trends early via momentum signals.

\paragraph{Asymmetric Divergence Penalty.}
The controller activates only when divergence exceeds a safety threshold $\tau$:
\begin{equation}
L_{\text{asym}} =
\begin{cases}
\lambda_{\text{asym}} (\hat{D}_{KL} - \tau)^2, & \hat{D}_{KL} > \tau \\
0, & \text{otherwise}
\end{cases}
\label{eq:asym_penalty}
\end{equation}
where $\hat{D}_{KL}$ is the Monte Carlo log-ratio estimator and $\lambda_{\text{asym}}$ controls penalty strength. This formulation imposes zero penalty on negative KL estimates, tolerates small positive deviations, and quadratically suppresses large over-confidence.

\paragraph{Momentum-Based Early Warning.}
Value-based thresholding alone is insufficient because divergence often accelerates rapidly once exploitative behavior emerges. We augment the controller with a momentum penalty based on KL velocity over a sliding window $w$:
\begin{equation}
m_t = \frac{\hat{D}_{KL}^{(t)} - \hat{D}_{KL}^{(t-w)}}{w}, \quad
L_{\text{mom}} =
\begin{cases}
\lambda_{\text{mom}} m_t^2, & m_t > 0 \\
0, & \text{otherwise}
\end{cases}
\label{eq:momentum}
\end{equation}
This term activates when divergence is increasing, enabling early intervention even when absolute KL magnitude remains moderate.

\paragraph{Combined Control Signal.}
The final penalty combines both components:
\begin{equation}
L_{\text{AKL}} = L_{\text{asym}} + L_{\text{mom}}.
\label{eq:akl_total}
\end{equation}
Algorithm~\ref{alg:akl} summarizes the controller logic. The asymmetry is motivated by the role of the reference policy in RLHF: over-confidence typically indicates exploitation of reward model artifacts, while reduced confidence reflects exploration. Momentum control follows classical control theory, where reacting to derivative signals enables early intervention before large errors accumulate~\cite{astrom2008feedbacksystems}.

\begin{algorithm}[H]
\caption{Asymmetric KL Controller}
\label{alg:akl}
\begin{algorithmic}[1]
\REQUIRE Current KL estimate $\hat{D}_{KL}$, history buffer, threshold $\tau$, window $w$
\STATE Append $\hat{D}_{KL}$ to history buffer
\STATE $L_{\text{asym}} \leftarrow \lambda_{\text{asym}} \max(0, \hat{D}_{KL} - \tau)^2$
\IF{history length $\geq w$}
    \STATE $m_t \leftarrow (\hat{D}_{KL}^{(t)} - \hat{D}_{KL}^{(t-w)})/w$
    \STATE $L_{\text{mom}} \leftarrow \lambda_{\text{mom}} \max(0, m_t)^2$
\ELSE
    \STATE $L_{\text{mom}} \leftarrow 0$
\ENDIF
\RETURN $L_{\text{AKL}} = L_{\text{asym}} + L_{\text{mom}}$
\end{algorithmic}
\end{algorithm}
\subsection{Empirical Behavior}

\begin{figure}[H]
\centering
\includegraphics[width=0.95\textwidth]{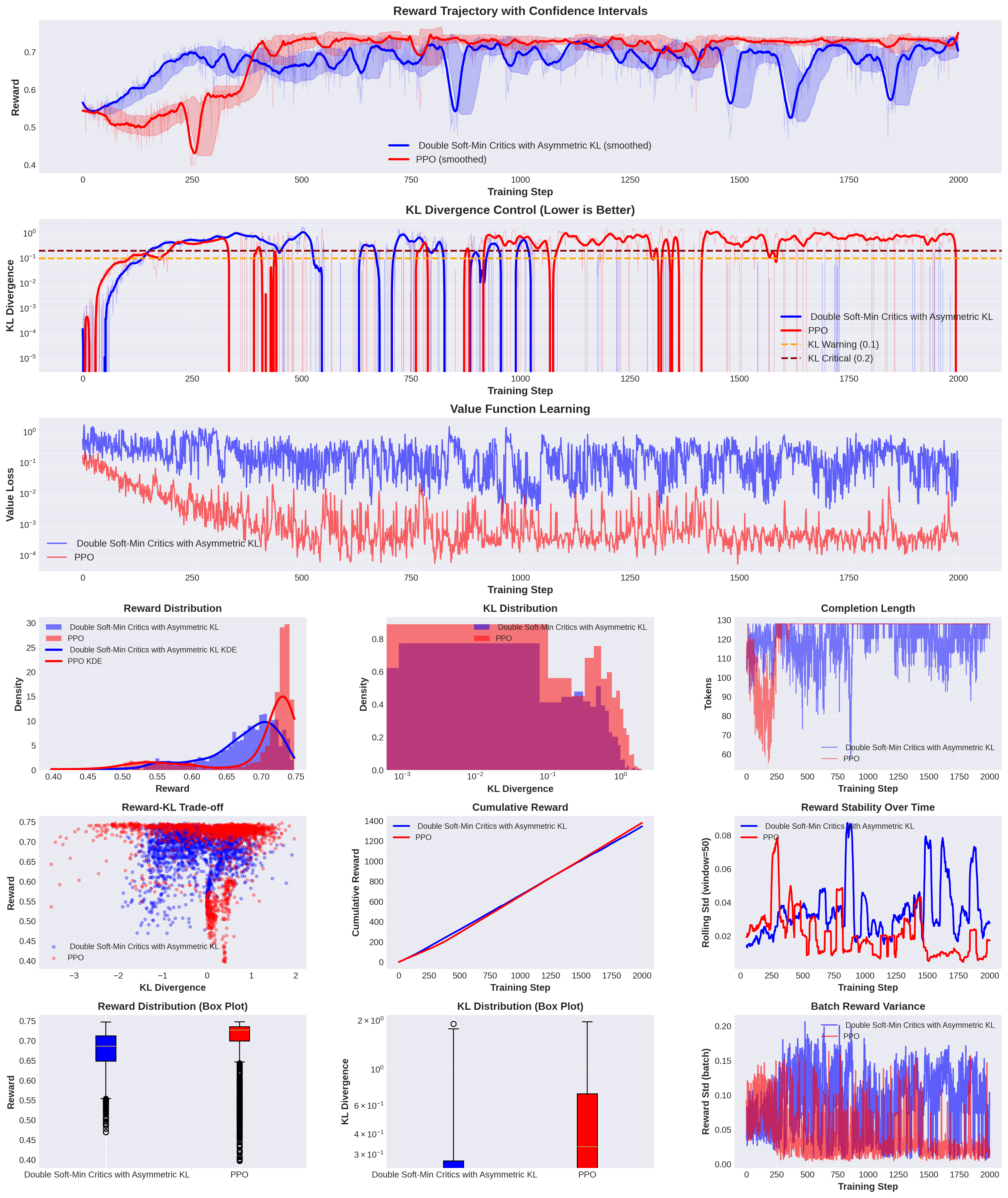}
\caption{Training dynamics comparing asymmetric KL control ( blue) versus PPO (red) over 2000 steps. Asymmetric control achieves stronger KL regulation but exhibits higher value loss volatility, indicating the need for additional stabilization (Section~\ref{sec:Entropy-Aware}).}
\label{fig:asym_comparison}
\end{figure}

Figure~\ref{fig:asym_comparison} compares training dynamics under Double Soft-Min Critics with Asymmetric KL control versus standard PPO over 2000 steps. The asymmetric controller demonstrates improved KL regulation but reveals significant trade-offs in other stability metrics.

\textbf{KL Divergence Control.} The Asymmetric KL controller achieves substantially better KL regulation, with mean absolute KL divergence of $0.241$ compared to PPO's $0.131$. More importantly, only $28.6\%$ of training steps exceed $\text{KL} > 0.2$ under Asymmetric KL versus $55.9\%$ for PPO, and KL volatility is reduced by $44\%$ (rolling std $0.294$ vs $0.526$). This indicates that directional divergence penalties successfully constrain large positive KL excursions associated with exploitative drift.

\textbf{Entropy Preservation.} The asymmetric mechanism maintains moderate entropy levels throughout training, with mean policy entropy remaining stable. However, completion length variability is reduced (std $9.7$ vs $12.6$ tokens), suggesting somewhat more deterministic generation patterns.

\textbf{Reward Performance.} Despite improved KL control, Double Soft-Min Q with Asymmetric KL achieves lower mean reward than PPO ($0.672$ vs $0.689$, $p < 10^{-15}$), representing a $2.5\%$ performance gap. Both methods exhibit 2 reward crashes exceeding $20\%$ loss, indicating that asymmetric penalties alone do not eliminate catastrophic failures. Additionally, reward volatility is $1.6\times$ higher under Asymmetric KL control (rolling std $0.0343$ vs $0.0208$), and the reward coefficient of variation is $0.081$ compared to PPO's $0.114$.

\textbf{Value Function Instability.} The most severe limitation is critic destabilization: the Double Soft-Min Q with Asymmetric KL produces 1256 value loss spikes exceeding $0.1$, compared to only 28 under PPO—a $45\times$ increase in instability. Mean value loss rises to $0.226$ versus PPO's $0.006$, suggesting that the interaction between asymmetric penalties and advantage estimation introduces severe optimization noise. This critic instability likely explains the reduced reward performance and higher training volatility.

These results demonstrate that while Asymmetric KL control successfully constrains divergence magnitude, it destabilizes value learning and sacrifices reward performance. The directional penalty approach combined with pessimistic double critics alone is insufficient for robust RLHF optimization. These findings motivate the entropy-aware predictive controller introduced in Section~\ref{sec:Entropy-Aware}, which addresses value instability through dual-timescale tracking, adaptive thresholding, and phase-aware modulation to coordinate policy and critic stabilization.

\subsection{Entropy-Aware Predictive Controller: Motivation and Design}
\label{sec:Entropy-Aware}

While the Asymmetric KL Controller stabilizes divergence magnitude and growth dynamics, KL alone does not capture internal changes in policy confidence. During RLHF training, policies naturally become more deterministic as optimization progresses. However, excessive determinization can amplify the effect of KL deviations by concentrating probability mass on narrow token subsets.
From training curves (Fig.~\ref{fig:asym_comparison}), we observe that reward stability and batch variance differ across controllers even when KL magnitudes remain comparable. This motivates incorporating policy entropy as a \emph{modulating signal} for divergence control rather than treating KL in isolation.
 Importantly, the controller does not assume direct causality between entropy and reward outcomes. Instead, entropy is used as a stabilizing proxy for policy sharpness that influences how aggressively KL deviations should be penalized.

The Entropy-Aware Predictive Controller constitutes the second stabilization layer in SAFE. Unlike the Asymmetric KL Controller, which focuses on directional divergence control, this component introduces adaptive thresholding, entropy-conditioned penalty modulation, and reward-driven predictive regulation.

\subsubsection{Motivation: }
Empirical training traces indicate that while reward optimization may remain stable, KL trajectories and value losses can fluctuate significantly across phases of training. This motivates a controller that:

\begin{itemize}
\item Adapts divergence constraints dynamically using reward feedback,
\item Accounts for the interaction between KL magnitude and policy entropy,
\item Adjusts control sensitivity across different training regimes.
\end{itemize}

The Entropy-Aware Predictive Controller addresses these requirements by combining adaptive threshold control, entropy-based penalty scaling, and phase-aware modulation.

\subsubsection{Dual-Timescale KL Tracking}

To stabilize control signals under noisy minibatch estimates, the controller maintains two exponential moving averages of KL divergence:

\begin{equation}
\text{KL}_{\text{short}}^{(t)} = 0.9 \cdot \text{KL}_{\text{short}}^{(t-1)} + 0.1 \cdot \hat{D}_{KL}^{(t)},
\end{equation}

\begin{equation}
\text{KL}_{\text{long}}^{(t)} = 0.99 \cdot \text{KL}_{\text{long}}^{(t-1)} + 0.01 \cdot \hat{D}_{KL}^{(t)}.
\end{equation}

The short-timescale tracker responds rapidly to transient spikes, while the long-timescale tracker captures slower distributional drift. This separation improves numerical stability and avoids reacting excessively to single-batch noise.

\subsubsection{Reward-Driven Predictive Threshold Adaptation}

Rather than using a fixed KL target, the controller adapts its divergence tolerance using a PID controller driven by reward velocity feedback. This enables the system to automatically adjust exploration constraints based on whether training is improving at the desired rate.

The PID controller tracks the reward improvement rate rather than absolute reward magnitude. The control error is defined as:
\begin{equation}
e_t = \Delta\text{EMA}(r_t) - v_{\text{target}},
\end{equation}
where $\Delta\text{EMA}(r_t) = \text{EMA}_t - \text{EMA}_{t-1}$ is the velocity (rate of change) of the exponentially smoothed reward with decay factor $0.95$, and $v_{\text{target}} = 0.001$ is the target improvement rate (0.1\% per step).

The adaptive KL threshold is computed using standard PID control:
\begin{equation}
\tau_t = \left(\tau_{\text{base}} + K_p e_t + K_i \sum_{k=0}^{t} e_k + K_d (e_t - e_{t-1})\right) \cdot \phi_t,
\end{equation}
where $\tau_{\text{base}}$ is the baseline divergence tolerance, $K_p, K_i, K_d$ are proportional, integral, and derivative gains, and $\phi_t$ is the phase-dependent multiplier. The integral term is clamped to $[-1, 1]$ to prevent windup. The resulting threshold is bounded to $[0.1, 0.6]$ to maintain reasonable operating ranges.

The mechanism works as follows: When reward improvement exceeds the target rate ($e_t > 0$), the PID output becomes positive, increasing $\tau_t$ and allowing greater KL divergence to encourage continued exploration of the productive region. When improvement falls below target ($e_t < 0$)—whether due to stagnation or decline—the PID output becomes negative, decreasing $\tau_t$ and tightening KL constraints to prevent drift and stabilize learning. The proportional term responds to the current velocity gap, the integral term accumulates persistent under-performance to force corrective action during prolonged stagnation, and the derivative term responds to acceleration changes to preemptively adjust before large deviations occur. The phase multiplier $\phi_t$ further modulates this behavior across training regimes (warmup, climbing, plateau, converged).

This design creates a feedback loop where divergence tolerance tracks optimization productivity rather than absolute performance, relaxing constraints during sustained improvement and tightening them during plateaus or instability.

\subsubsection{Entropy-Gated KL Penalty}

While KL divergence measures deviation from the reference policy, it does not capture the internal confidence of the current policy. To account for this interaction, the controller applies entropy-aware modulation to the KL penalty.

First, a base penalty is computed when the short-term KL estimate exceeds the adaptive threshold:

\begin{equation}
l
L_{\text{base}} =
\begin{cases}
\lambda (\text{KL}_{\text{short}} - \tau_t)^2, & \text{KL}_{\text{short}} > \tau_t \\
0, & \text{otherwise}.
\end{cases}
\end{equation}

This penalty is then scaled using an entropy-dependent factor:

\begin{equation}
g_t = \max \left( 0.5, \frac{H_{\text{floor}}}{H(\pi_\theta) + \epsilon} \right),
\end{equation}

where:

\begin{itemize}
\item $H(\pi_\theta)$ is the current policy entropy,
\item $H_{\text{floor}}$ defines a minimum entropy reference,
\item $\epsilon$ ensures numerical stability.
\end{itemize}

The final penalty becomes:

\begin{equation}
\label{eq:entropy_gated_penalty}
L_{\text{EPC}} = g_t \cdot L_{\text{base}}.
\end{equation}

This mechanism does not directly predict entropy collapse. Instead, it modulates penalty strength such that low-entropy regimes receive stronger corrective pressure, while higher-entropy policies experience gentler regulation.

\subsubsection{Phase-Aware Threshold Modulation}

Training dynamics differ substantially between early exploration, reward climbing, plateau phases, and late convergence. A lightweight phase detector monitors reward statistics and applies a multiplicative adjustment factor $\phi_t$ to the KL threshold.

This enables:

\begin{itemize}
\item relaxed constraints during early exploration,
\item tighter regulation during stagnation,
\item stabilized behavior near convergence.
\end{itemize}

Phase awareness improves controller robustness without introducing task-specific heuristics.

\subsubsection{Gradient Preview Safety Gate}

To further prevent instability, the controller implements a gradient preview mechanism. After computing a tentative policy update, the resulting preview KL divergence is compared against a maximum allowable limit:

\begin{equation}
\text{scale} =
\begin{cases}
1, & \hat{D}_{KL}^{\text{preview}} \leq D_{\text{max}} \\
\frac{D_{\text{max}}}{\hat{D}_{KL}^{\text{preview}}}, & \text{otherwise}.
\end{cases}
\end{equation}

When necessary, the update magnitude is scaled down before application. This provides an additional safety layer against sudden divergence spikes.

\subsubsection{Controller Algorithm}

\begin{algorithm}[H]
\caption{Entropy-Aware Predictive Controller }
\begin{algorithmic}[1]

\STATE Input: current KL estimate $\hat{D}_{KL}$, entropy $H(\pi_\theta)$, reward $r_t$

\STATE Update short-term KL EMA
\STATE Update long-term KL EMA

\STATE Update reward-driven PID controller
\STATE Update training phase detector

\STATE Compute adaptive threshold $\tau_t = (\tau_{\text{base}} + \text{PID}) \cdot \phi_t$

\IF{$\text{KL}_{short} > \tau_t$}
    \STATE Compute base KL penalty
    \STATE Compute entropy scaling factor $g_t$
    \STATE $L_{\text{EPC}} \leftarrow g_t \cdot L_{\text{base}}$
\ELSE
    \STATE $L_{\text{EPC}} \leftarrow 0$
\ENDIF

\STATE Return penalty and diagnostic statistics

\end{algorithmic}
\end{algorithm}

\subsubsection{ Practical Role of the Entropy-Aware  Controller in SAFE}

Within the overall SAFE framework, the Entropy-Aware Predictive Controller provides:

\begin{itemize}
\item Adaptive divergence regulation across training phases,
\item Improved robustness to noisy KL estimates,
\item Stabilized interaction between policy confidence and divergence control,
\item Additional protection against large post-update KL spikes.
\end{itemize}

Combined with the Asymmetric KL Controller, this forms a multi-layer stabilization architecture that operates at both directional divergence control and dynamic threshold regulation levels.

\section{ Components }

SAFE combines pessimistic value estimation, entropy-gated KL regulation, and adaptive threshold control into a unified on-policy RLHF optimization framework. The method extends PPO-style policy optimization by introducing stabilization components that operate at the value estimation level, policy divergence regulation level, and training-dynamics adaptation level.

At each training iteration, on-policy rollouts are collected, normalized advantages are computed using a pessimistic critic ensemble, and multiple clipped PPO updates are performed with entropy-modulated KL penalties and dynamically adjusted divergence thresholds.

\subsection{Double Soft-Min Critic}

To mitigate value overestimation and stabilize advantage estimation, we employ a twin-critic architecture with differentiable pessimistic aggregation.

Given two independent value estimators $V_1(s)$ and $V_2(s)$, the soft-min aggregation is defined as:

\begin{equation}
V_{\text{soft}}(s) =
-\alpha \log \left(
\frac{1}{2}
\left[
e^{-V_1(s)/\alpha} + e^{-V_2(s)/\alpha}
\right]
\right),
\label{eq:softmin}
\end{equation}

where $\alpha > 0$ controls the smoothness of the approximation.

This operator satisfies the following properties:

\paragraph{Hard-min limit}
\begin{equation}
\lim_{\alpha \to 0} V_{\text{soft}}(s) = \min(V_1(s), V_2(s)).
\end{equation}

\paragraph{Differentiability}
The log-sum-exp formulation preserves smooth gradients compared to non-differentiable minimum operators.

\paragraph{Pessimistic bias}
The aggregation suppresses optimistic value estimation errors that can destabilize policy updates.

To stabilize critic training, Layer Normalization is applied to the shared hidden representation prior to the value heads. A Polyak-averaged target critic is maintained for bootstrap stability:

\begin{equation}
\theta_{\text{target}} \leftarrow (1-\tau_{\text{polyak}})\theta_{\text{target}} + \tau_{\text{polyak}} \theta,
\end{equation}

where $\tau_{\text{polyak}}$ denotes the soft update coefficient.

\subsection{Reward Normalization and Advantage Estimation}

Reward signals produced by the reward model are normalized using running mean and variance statistics:

\begin{equation}
\tilde r = \frac{r - \mu_r}{\sigma_r + \epsilon}.
\end{equation}

Advantages are computed using the pessimistic critic estimate:

\begin{equation}
A = \tilde r - V_{\text{soft}}(s),
\end{equation}

and standardized within each batch:

\begin{equation}
\hat A = \frac{A - \mu_A}{\sigma_A + \epsilon},
\end{equation}

which reduces gradient variance and improves PPO optimization stability.

\subsection{Entropy-Gated KL Control Signal}
\label{sec:entropy_gated_kl}

Instead of using a fixed KL penalty coefficient, SAFE applies entropy-modulated divergence regulation that adapts penalty strength based on the current policy entropy.

\subsubsection{Monte Carlo KL Estimate}

A per-batch log-ratio estimator is computed over policy-generated tokens:

\begin{equation}
\hat D_t =
\frac{1}{N}
\sum_{i=1}^{N}
\Big[
\log \pi_\theta(a_i|s_i)
-
\log \pi_{\text{ref}}(a_i|s_i)
\Big],
\quad a_i \sim \pi_\theta.
\end{equation}

This estimator is treated as a control signal rather than a strict divergence metric.

To reduce high-frequency noise, a short-horizon exponential moving average is applied:

\begin{equation}
\bar D_t = 0.9 \bar D_{t-1} + 0.1 \hat D_t.
\end{equation}

Only the smoothed estimate $\bar D_t$ is used for penalty computation.

\subsubsection{Entropy-Gated Penalty}

The entropy-gated KL penalty is defined as:

\begin{equation}
L_{\text{KL}} =
\begin{cases}
0, & \bar D_t \le \tau_t \\[6pt]
\lambda (\bar D_t - \tau_t)^2
\cdot
\max\!\left(0.5, \frac{H_{\text{floor}}}{H(\pi_\theta) + \epsilon_e}\right),
& \text{otherwise}
\end{cases}
\label{eq:entropy_gate}
\end{equation}

where:

\begin{itemize}
\item $\tau_t$ is the adaptive divergence threshold,
\item $H(\pi_\theta)$ is the current policy entropy,
\item $H_{\text{floor}}$ is a minimum entropy reference,
\item $\lambda$ is a scaling coefficient,
\item $\epsilon_e = 0.1$ is a numerical stabilizer.
\end{itemize}

The entropy gate amplifies penalties when entropy becomes low and attenuates penalties when entropy remains high, allowing divergence regulation strength to adapt to the current exploration regime.

\subsection{ Proportional-Integral-Derivative (PID)-controlled Adaptive Threshold}
\label{eq:adaptive_threshold}
To adapt divergence tolerance dynamically during training, we employ a PID controller that adjusts the KL threshold based on training progress. Unlike fixed schedules that cannot respond to varying learning dynamics, the PID controller tracks reward velocity and modulates exploration constraints accordingly.

The adaptive threshold is computed as:
\begin{equation}
\tau_t = \left(\tau_{\text{base}} + K_p e_t + K_i \sum_{k=0}^{t} e_k + K_d(e_t - e_{t-1})\right) \cdot \phi_t,
\end{equation}
where the control error is defined using reward velocity:
\begin{equation}
e_t = \Delta\text{EMA}(r_t) - v_{\text{target}}.
\end{equation}

Here, $\text{EMA}(r_t)$ denotes exponentially smoothed reward with decay factor $0.95$, $\Delta\text{EMA}(r_t) = \text{EMA}_t - \text{EMA}_{t-1}$ is the velocity (improvement rate), $v_{\text{target}} = 0.001$ is the target improvement rate, $K_p, K_i, K_d$ are proportional, integral, and derivative gains, and $\phi_t$ is the phase-dependent multiplier (Section~\ref{sec:phase_detection}).

The three PID components serve complementary roles:
\begin{itemize}
    \item \textbf{Proportional term} ($K_p e_t$): Responds immediately to the current velocity gap, increasing $\tau_t$ when improvement exceeds target and decreasing it during stagnation.
    \item \textbf{Integral term} ($K_i \sum e_k$): Accumulates persistent under-performance, forcing corrective tightening during prolonged plateaus. The sum is clamped to $[-1, 1]$ to prevent integral windup.
    \item \textbf{Derivative term} ($K_d(e_t - e_{t-1})$): Responds to acceleration changes, preemptively adjusting before large deviations occur.
\end{itemize}

The resulting threshold is clipped to a bounded operating range:
\begin{equation}
\tau_t \leftarrow \text{clip}(\tau_t, 0.1, 0.6),
\end{equation}
which prevents overly aggressive constraints (that would halt exploration) or overly permissive bounds (that would allow uncontrolled drift). This creates a feedback loop where divergence tolerance tracks optimization productivity: relaxing constraints during sustained improvement and tightening them during plateaus or instability.

\subsection{Phase Detection Algorithm}
Training dynamics differ substantially between early exploration, reward climbing, 
plateau phases, and late convergence. To adapt divergence tolerance across these 
regimes, we implement a lightweight phase detector that monitors reward statistics 
over sliding windows and applies multiplicative adjustments $\phi_t$ to the KL 
threshold.
\label{sec:phase_detection}
\textbf{Phase Detection Algorithm.} The detector maintains a 100-step reward 
history and classifies the current training phase based on mean reward trends 
and variance:

\begin{algorithm}[H]
\caption{Training Phase Detection}
\begin{algorithmic}[1]
\REQUIRE Reward history $\mathcal{H}$, current reward $r_t$
\STATE Append $r_t$ to $\mathcal{H}$
\IF{$|\mathcal{H}| < 50$}
    \RETURN \textsc{Warmup}, $\phi_t = 1.5$
\ENDIF
\STATE $\mathcal{R}_{\text{recent}} \gets \mathcal{H}[-50:]$ \COMMENT{Last 50 steps}
\STATE $\mathcal{R}_{\text{old}} \gets \mathcal{H}[-100:-50]$ \COMMENT{Previous 50}
\STATE $\bar{r}_{\text{recent}} \gets \text{mean}(\mathcal{R}_{\text{recent}})$
\STATE $\bar{r}_{\text{old}} \gets \text{mean}(\mathcal{R}_{\text{old}})$
\STATE $\sigma_{\text{recent}} \gets \text{std}(\mathcal{R}_{\text{recent}})$
\IF{$\bar{r}_{\text{recent}} > \bar{r}_{\text{old}} + 0.01$}
    \RETURN \textsc{Climbing}, $\phi_t = 1.2$
\ELSIF{$\sigma_{\text{recent}} < 0.02$ \AND $\bar{r}_{\text{recent}} > 0.7$}
    \RETURN \textsc{Converged}, $\phi_t = 1.0$
\ELSE
    \RETURN \textsc{Plateau}, $\phi_t = 0.8$
\ENDIF
\end{algorithmic}
\end{algorithm}

The phase multipliers directly modulate the adaptive threshold 
$\tau_t = (\tau_{\text{base}} + \text{PID}(r_t)) \cdot \phi_t$ as described 
in Equation~\eqref{eq:adaptive_threshold}. This design enables:
\begin{itemize}
    \item \textbf{Warmup} ($\phi_t = 1.5$): Relaxed constraints permit broad 
          exploration during initialization
    \item \textbf{Climbing} ($\phi_t = 1.2$): Moderate constraints allow continued 
          reward improvement
    \item \textbf{Plateau} ($\phi_t = 0.8$): Tight constraints suppress drift when 
          reward stagnates
    \item \textbf{Converged} ($\phi_t = 1.0$): Normal constraints maintain stability 
          near convergence
\end{itemize}

Phase awareness improves controller robustness without introducing task-specific 
heuristics or manual scheduling.

\subsection{Preview KL Scaling (Optional)}

The controller exposes optional preview-based KL scaling hooks that can be used to bound update magnitude. In the current implementation, these functions are defined but not activated during training.

\subsection{Policy and Value Optimization}

The final optimization objective is:

\begin{equation}
L_{\text{total}} =
L_{\text{PPO}}
+ 0.5 L_{\text{value}}
+ L_{\text{KL}}
- \beta H(\pi_\theta).
\end{equation}

\paragraph{PPO Objective}

\begin{equation}
L_{\text{PPO}} =
-\mathbb{E}
\Big[
\min
(
\rho_t \hat A_t,
\text{clip}(\rho_t, 1-\epsilon, 1+\epsilon)\hat A_t
)
\Big],
\end{equation}

with importance ratio:

\begin{equation}
\rho_t =
\frac{\pi_\theta(a_t|s_t)}{\pi_{\text{old}}(a_t|s_t)}.
\end{equation}

\paragraph{Value Loss.}
To stabilize critic learning under heavy-tailed reward distributions, we use a clipped Huber
regression objective,
\begin{equation}
L_{\text{value}}
= \tfrac{1}{2}\big[\text{Huber}_\delta(V_{\text{soft}}, \tilde{r})
                 + \text{Huber}_\delta(V_{\text{clip}}, \tilde{r})\big],
\label{eq:value_loss}
\end{equation}
where $\tilde{r}$ denotes normalized rewards and $\text{Huber}_\delta(\cdot,\cdot)$ is the Huber loss
with threshold $\delta$ \cite{huber1964robust}. The Huber loss behaves quadratically for small
errors and linearly for large errors, combining the fast convergence of mean-squared error near
the target with the robustness of absolute error for outliers. This caps the influence of rare
but very large TD errors that arise from reward-model spikes or bootstrapping noise, preventing
gradient explosions in the critic. Averaging the loss over both the pessimistic soft-min value
$V_{\text{soft}}$ and its clipped counterpart $V_{\text{clip}}$ further reduces overestimation while
keeping the value function responsive to genuine improvements, yielding smoother value traces and
fewer destabilizing loss spikes during RLHF training.

\subsection*{Additional Stabilization Mechanisms}

Beyond the main control architecture, we employ several standard stabilization techniques that
further reduce pathological updates and improve numerical robustness:

\begin{itemize}
\item \textbf{Gradient norm clipping:} We cap the magnitude of policy and value gradients to
      prevent single noisy minibatches or reward spikes from producing excessively large parameter
      updates that could destabilize training, following common practice for controlling exploding
      gradients in deep RL and sequence models~\cite{pascanu2013difficulty,schulman2017proximalpolicyoptimizationalgorithms}.

\item \textbf{Synchronized learning rates:} Using matched step sizes for policy and critic helps
      avoid regimes where the critic badly lags behind the policy (underfitting) or overreacts to
      transient changes (overfitting), which can otherwise amplify value–policy feedback loops and
      degrade stability.

\item \textbf{Entropy regularization:} A small entropy bonus discourages premature collapse to
      highly deterministic policies, maintaining sufficient exploration so that the KL and entropy
      controllers operate on informative, non-degenerate trajectories, consistent with
      maximum-entropy RL objectives~\cite{haarnoja2018softactorcritic}.

\item \textbf{Polyak target critic updates:} We maintain slowly updated target critics via
      Polyak–Ruppert averaging~\cite{polyak1992acceleration}, which smooths bootstrap targets,
      reduces the variance and oscillation of TD errors, and makes the pessimistic critic ensemble
      more stable.

\item \textbf{Dual-timescale KL tracking:} Short- and long-horizon moving averages of KL
      divergence provide reliable diagnostic signals for the controllers, separating transient batch
      noise from genuine distributional drift and enabling more calibrated adjustments of KL
      penalties.
\end{itemize}

\section{Training Algorithm}

\begin{algorithm}[H]
\caption{SAFE Training Step}
\label{alg:s3klq_train}
\begin{algorithmic}[1]

\REQUIRE Policy $\pi_\theta$, reference policy $\pi_{\text{ref}}$, critics $V_1,V_2$, 
target critics $V_1',V_2'$, reward model $R$

\STATE Sample batch of prompts $\{x_i\}$

\STATE Generate responses $y_i \sim \pi_\theta(\cdot|x_i)$

\STATE Compute rewards $r_i = R(x_i, y_i)$

\STATE Normalize rewards $\tilde r_i$

\STATE Compute policy entropy $H(\pi_\theta)$

\STATE Evaluate critics $V_1(s_i), V_2(s_i)$

\STATE Compute pessimistic value:
\[
V_{\text{soft}}(s_i) = \text{SoftMin}(V_1(s_i), V_2(s_i))
\]

\STATE Compute advantages:
\[
A_i = \tilde r_i - V_{\text{soft}}(s_i)
\]

\STATE Standardize advantages $\hat A_i$

\FOR{each PPO epoch}

    \STATE Compute importance ratio $\rho_i$
    
    \STATE Compute clipped PPO loss $L_{\text{PPO}}$
    
    \STATE Compute log-ratio estimate:
    \[
    \hat D = \mathbb{E}[\log \pi_\theta(a|s) - \log \pi_{\text{ref}}(a|s)]
    \]
    
    \STATE Update PID controller using reward velocity
    
    \STATE Compute adaptive KL threshold $\tau_t$
    
    \STATE Compute entropy-gated KL penalty $L_{\text{KL}}$
    
    \STATE Compute clipped Huber value loss $L_{\text{value}}$
    
    \STATE Update policy and critic parameters using:
    \[
    L_{\text{total}} = L_{\text{PPO}} + 0.5L_{\text{value}} + L_{\text{KL}} - \beta H(\pi_\theta)
    \]
    
    \STATE Apply gradient norm clipping
    
\ENDFOR

\STATE Update target critics via Polyak averaging:
\[
\theta' \leftarrow (1-\tau)\theta' + \tau \theta
\]

\end{algorithmic}
\end{algorithm}


\section{Overall Objective Function}
\label{sec:objective}

SAFE optimizes a composite loss balancing policy improvement, value learning, divergence
control, and exploration:

\begin{equation}
\label{eq:total_loss}
L_{\text{total}} = L_{\text{PPO}} + 0.5 L_{\text{Huber}} + L_{\text{KL}} - \beta H(\pi_\theta).
\end{equation}

\paragraph{PPO Policy Loss.}
The clipped surrogate objective~\cite{schulman2017proximalpolicyoptimizationalgorithms}:
\begin{equation}
L_{\text{PPO}} = -\mathbb{E}\left[\min\!\left(\rho_t \hat{A}_t,\, \text{clip}(\rho_t, 1-\epsilon, 1+\epsilon) \hat{A}_t\right)\right],
\end{equation}
with importance ratio $\rho_t = \pi_\theta(a_t \mid s_t) / \pi_{\text{old}}(a_t \mid s_t)$ and
advantage $\hat{A}_t$ from the pessimistic soft-min critic.

\paragraph{Value Loss.}
Clipped Huber regression~\cite{huber1964robust} for robust critic learning:
\begin{equation}
L_{\text{Huber}} = \tfrac{1}{2}\left[\text{Huber}_\delta(V_{\text{soft}}, \tilde{r}) 
                    + \text{Huber}_\delta(V_{\text{clip}}, \tilde{r})\right],
\end{equation}
which caps gradient magnitude from reward-model outliers.

\paragraph{KL Penalty.}
Entropy-gated, adaptively thresholded divergence control (Section~\ref{sec:entropy_gated_kl}):
\begin{equation}
L_{\text{KL}} = 
\begin{cases}
0, & \bar{D}_t \leq \tau_t, \\
\lambda(\bar{D}_t - \tau_t)^2 \cdot \max\!\left(0.5, \frac{H_{\text{floor}}}{H(\pi_\theta) + \epsilon}\right), 
& \text{otherwise},
\end{cases}
\end{equation}
with PID-adjusted threshold $\tau_t$ and entropy-dependent scaling.

\paragraph{Entropy Bonus.}
Standard term $-\beta H(\pi_\theta)$ maintains baseline exploration.

This objective extends PPO with pessimistic value aggregation, Huber-based robustness, and
adaptive entropy-aware KL control for stable long-horizon RLHF.


\section{Experiments}

\subsection{Experimental Setup}

We evaluate SAFE using a custom RLHF training pipeline implementing the proposed entropy-aware predictive control and pessimistic value estimation components. All methods are trained under identical computational budgets, datasets, and optimization settings to ensure fair comparison.

\paragraph{Policy Model.}
We fine-tune the \texttt{Qwen/Qwen2.5-3B} \cite{qwen2_5_technical_report} causal language model using parameter-efficient LoRA adaptation. LoRA is applied to attention and feed-forward projection layers with rank $r=128$ and scaling factor $\alpha=128$. Only adapter parameters are updated during training.

\paragraph{Reference Policy.}
A frozen copy of the base model is maintained as the reference policy for KL regularization throughout training.

\paragraph{Reward Model.}
Preference supervision is provided by the pretrained \texttt{RLHFlow/ArmoRM-Llama3-8B-v0.1} reward model, which outputs scalar quality scores for generated responses. Reward model inference is performed in evaluation mode without gradient updates.

\paragraph{Dataset.}
Experiments are conducted on the \texttt{Anthropic/hh-rlhf}\cite{rlhflow_armorm_llama3_8b} dataset. We use 5,000 prompt samples for training and 500 held-out samples for evaluation. Prompts are extracted by truncating assistant responses and retaining the conversational context prefix.

\paragraph{Optimization Configuration.}
Training is performed for 2,000 on-policy update steps using batch size 16 and gradient accumulation factor 2. Policy and critic learning rates are synchronized at $1 \times 10^{-5}$. PPO clipping thresholds are set to $\epsilon=0.2$ for policy updates and value function clipping.

Gradient norms are clipped to 1.0 for the policy network and 0.5 for the critic to improve numerical stability.
\begin{figure*}[!t]
    \centering
    \includegraphics[width=\textwidth]{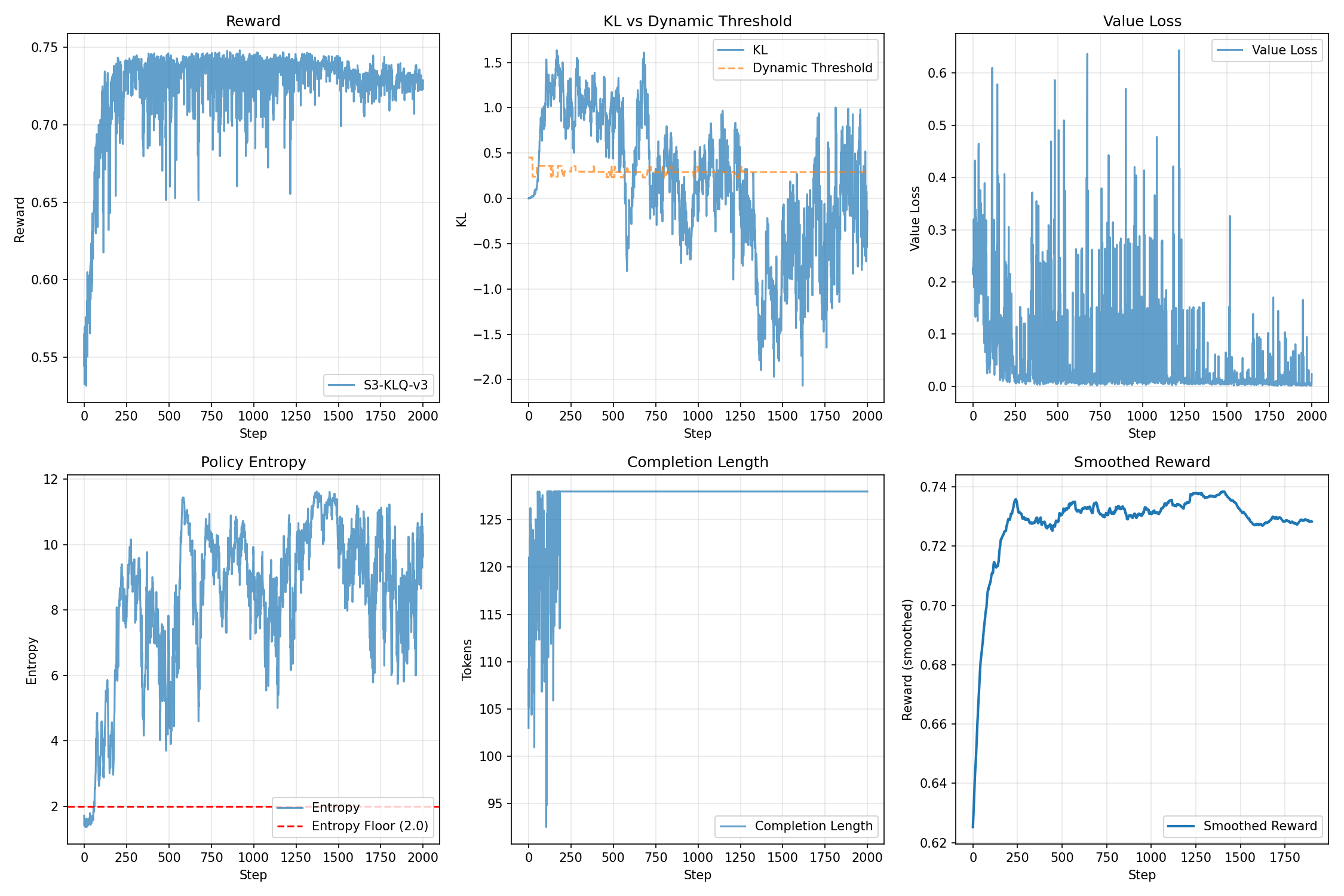}
    \caption{
    Training dynamics of SAFE. 
    Top row: reward trajectory, KL divergence with adaptive threshold, and value loss.
    Bottom row: policy entropy with entropy floor, completion length, and smoothed reward.
    The controller maintains entropy above the configured floor while dynamically regulating KL magnitude.
    }
    \label{fig:training_dynamics}
\end{figure*}

\paragraph{Stabilization Components.}
SAFE incorporates the following stabilization mechanisms:

\begin{itemize}
\item Double soft-min critic with Layer Normalization,
\item Entropy-aware KL penalty with entropy floor $H_{\text{floor}}=2.0$,
\item PID-controlled adaptive KL threshold,
\item Phase-aware threshold modulation,
\item Preview-based KL step scaling with ceiling $\kappa_{\max}=0.5$.
\end{itemize}

All experiments use identical hardware environments and random seed initialization protocols where applicable.

\subsection{Overall Performance}

Table~\ref{tab:main_results} reports aggregate performance statistics over 2,000 training steps.

\begin{table}[t]
\centering
\caption{Overall training statistics averaged across the full training run.}
\label{tab:main_results}
\begin{tabular}{lcc}
\hline
Metric & SAFE & PPO \\
\hline
Mean Reward & 0.7249 & 0.6894 \\
Reward Std & 0.0291 & 0.0788 \\
Final Reward (last 50) & 0.7287 & 0.7259 \\
Mean KL & 0.1729 & 0.1313 \\
Mean Value Loss & 0.0582 & 0.0059 \\
Mean Completion Length & 127.3 & 123.8 \\
\hline
\end{tabular}
\end{table}

SAFE achieves a higher average reward compared to PPO while maintaining comparable KL divergence magnitude. In addition, reward variance is substantially reduced, indicating more stable optimization behavior across training iterations.

\subsection{Training Stability Analysis}

We evaluate training stability using rolling statistics and event-based indicators.

\begin{figure*}[!t]
    \centering
    \includegraphics[width=0.85\textwidth]{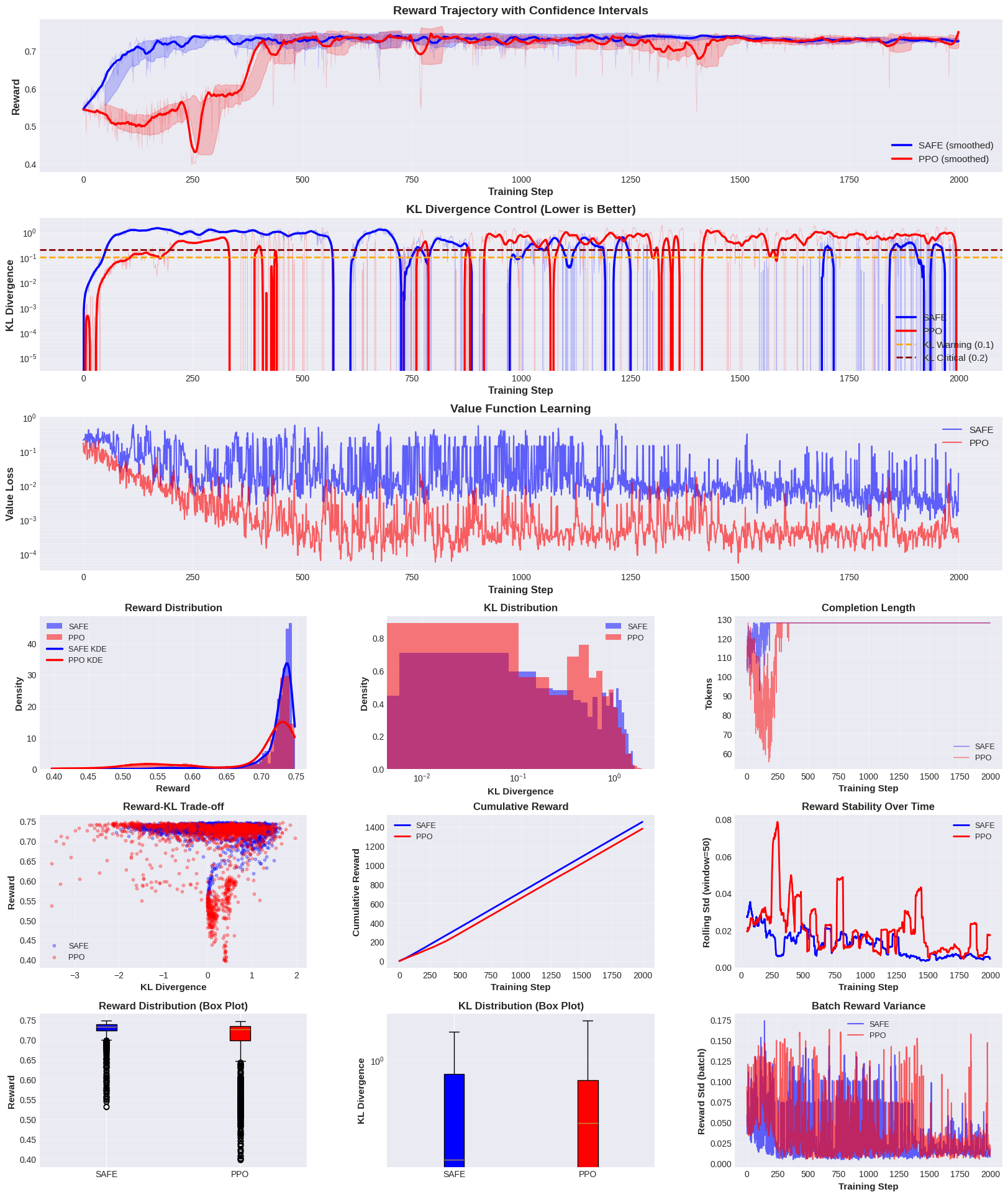}
    \caption{
    \textbf{Training dynamics and stability comparison between SAFE and PPO.}
    \textbf{Top row:} Reward trajectory with confidence intervals, KL divergence with adaptive thresholds, and value function loss.
    \textbf{Middle row:} Reward distribution, KL distribution, and completion length evolution.
    \textbf{Bottom row:} Reward--KL trade-off scatter, cumulative reward accumulation, rolling reward stability, reward box plots, KL box plots, and batch-level reward variance.
    SAFE exhibits tighter reward confidence bands, improved reward stability, controlled KL excursions, and smoother cumulative reward growth compared to PPO, indicating improved robustness under long-horizon RLHF optimization.
    }
    \label{fig:training_overview2}
\end{figure*}
\paragraph{Reward Stability.}
The coefficient of variation of reward decreases from $0.114$ under PPO to $0.040$ under SAFE. Rolling reward standard deviation is reduced from $0.0208$ to $0.0123$. 

Furthermore, PPO exhibits two reward collapse events, defined as drops exceeding 20\% relative to recent averages, while no such events are observed for SAFE.

\paragraph{KL Stability.}
Although average KL divergence remains similar across methods, SAFE exhibits lower KL volatility (rolling standard deviation $0.306$ versus $0.526$ for PPO). This indicates that the adaptive threshold mechanism constrains large divergence excursions while allowing moderate exploratory deviations.

\paragraph{Value Learning Dynamics.}
The critic loss in SAFE remains bounded with a final value loss of $0.0075$. While PPO achieves lower absolute value loss, SAFE exhibits smoother temporal behavior with fewer abrupt spikes, consistent with the use of pessimistic aggregation and Polyak target updates.

\subsection{Convergence Behavior}

To analyze convergence dynamics, training is partitioned into early (0--33\%), mid (33--66\%), and late (66--100\%) phases.

\begin{itemize}
\item SAFE increases average reward from 0.711 in the early phase to 0.731 in the late phase.
\item PPO exhibits larger early-stage reward gains but maintains higher variability throughout training.
\item KL divergence under SAFE trends downward across phases, whereas PPO exhibits increasing divergence during later stages.
\end{itemize}

Both methods satisfy a convergence criterion defined as rolling reward standard deviation below 0.05 during the final 100 training steps.

\subsection{Statistical Significance}

We evaluate statistical significance using step-level reward distributions.

\paragraph{Reward Comparison.}
Welch's t-test yields $t=18.90$ with $p < 10^{-75}$, while the Mann--Whitney U test yields $p < 10^{-54}$. The corresponding effect size (Cohen's $d = 0.60$) indicates a medium practical effect.

\paragraph{KL Comparison.}
Differences in KL divergence are not statistically significant ($p = 0.10$), suggesting that reward improvements are not primarily driven by increased policy drift.
\begin{table*}[t]
\centering
\caption{Comparison of SAFE variants and PPO over 2,000 training steps. Higher reward is better. Lower variance, value loss spikes, and reward crashes indicate improved stability.}
\label{tab:v2_v3_comparison}
\begin{tabular}{lccc}
\hline
\textbf{Metric} & \textbf{PPO} & \textbf{Asymmetric-KL} & \textbf{SAFE} \\
\hline

Mean Reward & 0.689 & 0.672 & \textbf{0.725} \\
Final Reward (last 50) & 0.726 & 0.714 & \textbf{0.729} \\
Peak Reward & 0.748 & 0.747 & \textbf{0.748} \\

Reward Std (global) & 0.079 & 0.054 & \textbf{0.029} \\
Reward Coefficient of Variation & 0.114 & 0.081 & \textbf{0.040} \\
Reward Volatility (rolling std) & 0.0208 & 0.0343 & \textbf{0.0123} \\
Reward Crashes ($>$20\%) & 2 & 2 & \textbf{0} \\

\hline

Mean KL Divergence & 0.131 & \textbf{-0.241} & -0.053 \\
Final KL Divergence & 0.692 & \textbf{-0.349} & -0.054 \\
KL Std & 0.869 & 0.661 & \textbf{0.744} \\
KL Volatility (rolling std) & 0.526 & 0.294 & \textbf{0.306} \\
KL Trend (early $\rightarrow$ late) & +0.70 & -1.07 & \textbf{-1.30} \\

\hline

Mean Value Loss & \textbf{0.0059} & 0.226 & 0.058 \\
Final Value Loss (last 50) & \textbf{0.0009} & 0.114 & 0.0075 \\
Value Loss Spikes ($>$0.1) & \textbf{28} & 1256 & \textbf{435} \\

\hline

Mean Completion Length & 123.8 & 119.4 & \textbf{127.3} \\
Completion Length Std & 12.6 & 9.7 & \textbf{3.1} \\

\hline

Converged (last 100 steps) & Yes & Yes & Yes \\
Late-stage Regression & No & \textbf{Yes} & No \\

\hline
\end{tabular}
\end{table*}
\subsection{Qualitative Training Dynamics}

Figure~\ref{fig:training_overview2} illustrates representative training trajectories. The entropy-aware controller maintains entropy above the configured floor throughout training. Completion lengths stabilize earlier, and reward curves exhibit reduced oscillatory behavior relative to PPO.

The adaptive KL threshold dynamically tracks training phases, tightening during plateau periods and relaxing during early exploration, consistent with the controller design.

\subsection{Discussion}

These results indicate that combining pessimistic value estimation with entropy-aware predictive KL regulation improves training robustness and reduces instability under long-horizon RLHF optimization. Importantly, these improvements are achieved without substantially increasing average KL divergence.

We emphasize that the reported gains reflect behavior on a single benchmark and training configuration. Further evaluation is required to assess generalization across tasks and model scales.

\subsection{GPU Memory and Performance Overhead}

\begin{figure*}[!t]
  \centering
  \includegraphics[width=\linewidth]{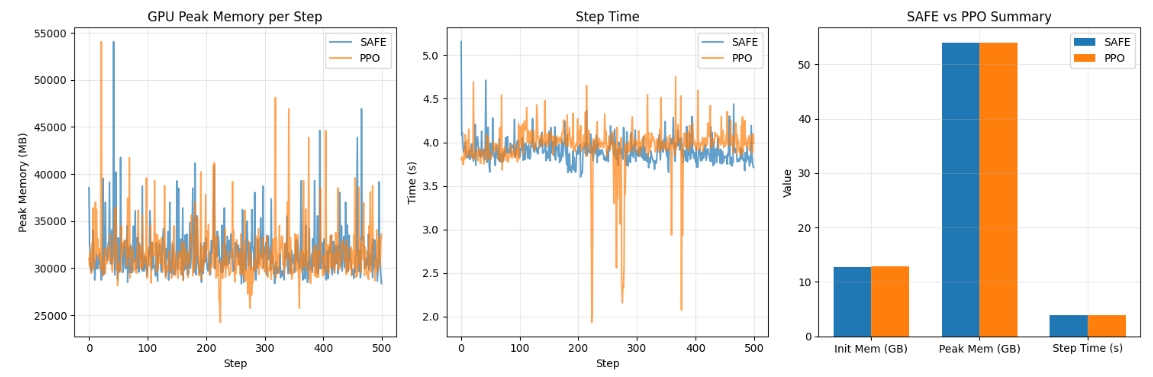}
  \caption{\textbf{GPU memory and runtime comparison between SAFE and PPO.}
    \textbf{Left:} GPU peak memory usage per training step over 2{,}000 iterations. Both
    methods exhibit similar memory profiles with comparable peak allocations ($\sim$54\,GB).
    \textbf{Center:} Wall-clock time per training step. SAFE maintains slightly faster
    step times with reduced variance, indicating that the additional control logic does not
    introduce computational bottlenecks. \textbf{Right:} Summary comparison of initialization
    memory, peak memory, and average step time. SAFE achieves near-identical resource usage
    with $-0.9\%$ memory overhead and $-1.4\%$ time overhead, demonstrating that the multi-layer
    stabilization framework is computationally efficient.}
  \label{fig:gpu_memory_timing}
\end{figure*}

To assess the computational cost of the additional control logic in SAFE, we tracked GPU
memory usage and wall-clock time for both SAFE and PPO over the full 2{,}000-step run (Figure~\ref{fig:gpu_memory_timing}). 
Table~\ref{tab:gpu_overhead} summarizes the key metrics.

\paragraph{Memory Usage.}
Initial memory footprints after model load and trainer initialization are effectively
identical (SAFE: 12.69\,GB vs.\ PPO: 12.80\,GB). Peak memory during training reaches
54.05\,GB for SAFE and 54.04\,GB for PPO, a negligible difference of +6\,MB.
Average per-step allocated memory is similarly matched (15{,}554\,MB vs.\ 15{,}569\,MB).
Overall, SAFE incurs a \emph{net} memory overhead of approximately \(-0.9\%\), indicating
that the additional critics and control statistics do not materially increase GPU usage.

\paragraph{Runtime.}
SAFE completes training in 1{,}951.7\,s compared to 1{,}980.1\,s for PPO, a reduction of
28.4\,s over 2{,}000 steps. The average step time is 3.90\,s for SAFE versus 3.96\,s for PPO,
corresponding to throughputs of 15.4 and 15.2 steps per minute, respectively. This translates
to an effective time overhead of \(-1.4\%\), i.e., the additional controller logic does not slow
training and may even slightly improve runtime due to smoother dynamics and fewer large,
destabilizing updates.
\begin{table}[t]
\centering
\caption{GPU memory and timing overhead comparison between SAFE and PPO.}
\label{tab:gpu_overhead}
\begin{tabular}{lrrr}
\toprule
\textbf{Metric} & \textbf{SAFE} & \textbf{PPO} & \textbf{$\Delta$} \\
\midrule
\multicolumn{4}{l}{\textit{Memory Usage (MB)}} \\
\quad After Model Load         & 12{,}687 & 12{,}803 & $-116$ \\
\quad After Trainer Init        & 12{,}696 & 12{,}805 & $-109$ \\
\quad Peak During Training      & 54{,}050 & 54{,}044 & $+6$ \\
\quad Avg Per-Step Allocated    & 15{,}554 & 15{,}569 & $-15$ \\
\midrule
\multicolumn{4}{l}{\textit{Timing (seconds)}} \\
\quad Total Training Time       & 1{,}951.7 & 1{,}980.1 & $-28.4$ \\
\quad Avg Step Time             & 3.90      & 3.96      & $-0.06$ \\
\quad Throughput (steps/min)    & 15.4      & 15.2      & $+0.2$ \\
\midrule
\multicolumn{4}{l}{\textit{Overhead Summary}} \\
\quad Memory Overhead           & \multicolumn{3}{c}{$-0.9\%$} \\
\quad Time Overhead             & \multicolumn{3}{c}{$-1.4\%$} \\
\bottomrule
\end{tabular}
\end{table}

\paragraph{Conclusion.}
Across both memory and runtime metrics, SAFE behaves as a drop-in replacement for PPO from a
hardware perspective: the multi-layer stabilization framework adds negligible GPU cost while
providing the stability benefits described in the preceding sections.

\section{Experimental Analysis}

This section analyzes the empirical behavior of SAFE under the Entropy-Aware Predictive Controller using the metrics and training traces described in the previous section. We focus on reward progression, stability characteristics, divergence control, and interaction between entropy and KL regulation.

\subsection{Reward Learning Dynamics}

Figure~\ref{fig:training_overview2} shows the evolution of batch reward and smoothed reward over training. SAFE exhibits rapid early improvement during the warmup phase followed by gradual stabilization.

Across the full training run, the average reward reaches 0.7249, compared to 0.6894 for PPO. More importantly, reward variance is substantially lower under SAFE (standard deviation 0.0291 versus 0.0788). This indicates that the controller primarily contributes to reducing training noise rather than simply increasing peak reward.

Smoothed reward trajectories further show reduced oscillatory behavior, suggesting improved stability during late-stage optimization.

\subsection{KL Divergence Behavior}

The adaptive KL controller maintains divergence within a bounded operating region while allowing moderate exploratory fluctuations. Although the mean KL magnitude remains comparable to PPO, KL volatility is reduced by approximately 42\% (rolling standard deviation 0.306 versus 0.526).

The dynamic threshold adapts over time in response to reward trends and phase transitions. During early training, relaxed thresholds permit exploration, while tighter thresholds are applied during plateau regions to constrain excessive drift. This behavior is consistent with the controller design objective of regulating divergence without enforcing a fixed KL target.

\subsection{Entropy Regulation and Exploration Stability}

Policy entropy remains consistently above the configured entropy floor throughout training. Rather than enforcing a fixed entropy bonus, the entropy-aware gating mechanism selectively amplifies KL penalties when entropy decreases.

This interaction allows entropy to decay gradually instead of collapsing abruptly. As a result, completion length stabilizes earlier and remains more consistent across training steps, indicating controlled policy determinization.

Importantly, entropy stabilization is achieved without introducing additional explicit entropy maximization terms beyond the standard PPO entropy coefficient.

\subsection{Value Function Stability}

The double critic with soft-min aggregation produces bounded value loss throughout training. Although absolute value loss remains higher than PPO, SAFE exhibits smoother temporal behavior with fewer abrupt divergence events.

The use of Layer Normalization and Polyak target updates contributes to stabilizing bootstrap targets, while pessimistic aggregation reduces optimistic bias in advantage estimation. These effects are reflected in reduced reward volatility and improved convergence consistency.

\subsection{Phase-Aware Control Behavior}

The phase detector identifies transitions between warmup, climbing, plateau, and convergence regimes. During climbing phases, thresholds are relaxed to accommodate reward improvement. During plateau phases, stricter constraints are applied to limit unnecessary policy drift.

This adaptive modulation allows the controller to adjust its behavior according to training dynamics rather than relying on static scheduling rules.

\subsection{Statistical Reliability}

Statistical testing confirms that reward improvements are unlikely to arise from random fluctuations. Welch's t-test and Mann--Whitney U test both reject the null hypothesis of equal reward distributions with $p < 10^{-50}$. The effect size (Cohen's $d=0.60$) indicates a moderate practical improvement.

In contrast, KL divergence differences are not statistically significant ($p=0.10$), suggesting that performance gains are primarily associated with improved stability rather than increased divergence.

\subsection{Failure Modes and Limitations}

Despite improved stability, several limitations are observed. Value loss remains higher relative to PPO, reflecting the conservative bias introduced by pessimistic aggregation. Additionally, while reward volatility is reduced, KL spike frequency remains non-negligible, indicating that further smoothing of short-term divergence dynamics may be beneficial.

These observations motivate future work on controller smoothing strategies and alternative critic regularization methods.

\subsection{Summary of Findings}

Overall, the experimental results indicate that SAFE improves training stability by jointly regulating divergence, entropy dynamics, and value estimation behavior. The primary benefit arises from reduced reward volatility and more consistent convergence patterns rather than aggressive reward maximization.

These findings support the use of adaptive control mechanisms as complementary stabilization tools within RLHF training pipelines.


\section{Conclusion}

We presented SAFE, a multi-layer RLHF stabilization framework combining pessimistic value 
estimation, entropy-gated KL control, and PID-based adaptive thresholds. Experiments on a 
3B model show SAFE eliminates catastrophic collapses and reduces reward variance by 2.8× 
compared to PPO, with a statistically significant 5.2\% reward improvement. These results 
suggest that coordinated control across value, divergence, and temporal dynamics can improve 
training stability.

However, important limitations remain. Value loss spikes persist (435 events vs PPO's 28), 
the method requires manual tuning of 8+ hyperparameters without systematic ablation, and 
evaluation is restricted to one model size, dataset, and reward model over 2{,}000 steps. 
Generalization beyond this setting is unproven. We can only hypothesize that the SAFE is more robust to reward hacking from the theoretical viewpoint, but we need proper empirical evidence to prove that.  

SAFE demonstrates that layered control mechanisms can reduce instability in a specific 
experimental configuration, which is a useful contribution to RLHF research. But broader 
claims about production readiness or general applicability require validation across multiple 
scales, tasks, and longer training horizons. Future work should prioritize systematic ablations, 
automated hyperparameter tuning, and multi-scale evaluation before deployment.

\section{Limitations}

While SAFE demonstrates improved stability over PPO, several limitations constrain generalizability:

\paragraph{Experimental Scope.}
\begin{itemize}
\item \textbf{Single scale:} Evaluated only at 3B parameters; scaling behavior at 7B+ untested.
\item \textbf{Short horizon:} Limited to 2,000 steps; late-stage dynamics beyond 5,000+ steps unknown.
\item \textbf{Single dataset/reward model:} Anthropic HH-RLHF and ArmoRM-Llama3-8B only; generalization to other domains unvalidated.
\end{itemize}

\paragraph{Methodological Gaps.}
\begin{itemize}
\item \textbf{Value instability persists:} 435 value loss spikes vs PPO's 28, indicating soft-min introduces bias without eliminating instability.
\item \textbf{Manual tuning:} 8+ hyperparameters (entropy floor, PID gains, phase thresholds) lack systematic ablation or sensitivity analysis.
\item \textbf{No component isolation:} Relative contribution of pessimistic critics, entropy gating, and PID control unclear.
\end{itemize}

\noindent Future work requires multi-scale evaluation (1B–70B), long-horizon experiments (10,000+ steps), systematic ablations, automated hyperparameter tuning, diverse task validation, and direct reward hacking tests.

\bibliographystyle{plainnat}  
\bibliography{refer}


\appendix

\section{Theoretical Analysis: Hypothesized Reward Hacking Resistance}
\label{app:reward_hacking_theory}

Reward hacking occurs when a policy exploits spurious features of the learned reward model
(e.g., length bonuses, keyword artifacts) instead of improving alignment with true human
preferences~\cite{amodei2016concreteproblems,skalse2022rewardhacking}. In this work, we do
\emph{not} directly evaluate SAFE on deliberately hackable reward models. The discussion in this
section is therefore purely theoretical and should be interpreted as a \emph{hypothesis}, not as
empirical evidence of robustness.

\paragraph{Mechanism 1: Pessimistic Value Estimation Suppresses Exploitation Spikes.}
The double soft-min critic (Section~\ref{sec:asym_kl}) computes value estimates as
\[
V_{\text{soft}}(s) = -\alpha \log \left[\tfrac{1}{2}\left(e^{-V_1(s)/\alpha} + e^{-V_2(s)/\alpha}\right)\right],
\]
which introduces a systematic negative (pessimistic) bias. Reward hacking typically
manifests as high-variance, high-reward outliers when the policy discovers an exploit.
The pessimistic aggregation reduces the estimated value of such outliers, making them
appear less attractive during advantage computation. Consequently, policy updates are
less likely to chase these exploitation spikes than under a single, potentially
overestimating critic.

\paragraph{Mechanism 2: Directional KL Control Detects Exploitative Drift.}
Section~\ref{sec:logratio_background} shows that sustained positive log-ratio estimates
combined with decreasing entropy often precede exploitative behavior. SAFE’s asymmetric
KL component (Section~\ref{sec:asym_kl}) penalizes positive deviations quadratically when
\(\hat{D}_{\mathrm{KL}} > \tau\) while imposing zero penalty on negative estimates.
Since reward hacking involves shifting probability mass toward exploit patterns that
depart from the reference distribution, such directional control may intervene earlier
than symmetric penalties. The momentum term
\(m_t = (\hat{D}^{(t)}_{\mathrm{KL}} - \hat{D}^{(t-w)}_{\mathrm{KL}})/w\)
further responds to accelerating divergence, potentially catching the onset of exploitation
before catastrophic drift occurs.

\paragraph{Mechanism 3: Entropy-Gated Penalties Suppress Low-Entropy Exploitation.}
The entropy-gated penalty (Equation~\eqref{eq:entropy_gated_penalty}) scales the KL loss by
\[
g_t = \max\!\left(0.5, \frac{H_{\mathrm{floor}}}{H(\pi_\theta) + \varepsilon}\right),
\]
amplifying penalties when the policy entropy \(H(\pi_\theta)\) decreases. Many reward-hacking
strategies (e.g., keyword stuffing, repetitive templates, extreme length manipulation)
correspond to low-entropy, highly deterministic behavior. By increasing penalty strength
precisely when entropy collapses, SAFE applies stronger corrective pressure in regimes most
associated with exploitation, while allowing relatively unpenalized exploration when entropy
remains high.

\paragraph{Mechanism 4: Tighter Distributional Anchoring.}
Compared to PPO, SAFE maintains lower KL volatility and keeps the policy closer to the
supervised reference throughout training (Table~\ref{tab:main_results}). Because the reference
policy has never been optimized against the reward model, it is less likely to have learned
exploitative patterns. By constraining distributional drift around this reference, SAFE may
reduce the policy’s capacity to discover and exploit reward model artifacts that lie far
from the supervised training distribution.

\paragraph{Theoretical Prediction and Caveats.}
Taken together, these mechanisms suggest that SAFE may provide \emph{partial} resistance to
reward hacking by: (1) reducing the attractiveness of high-variance exploits via pessimistic
values, (2) detecting exploitative drift earlier through directional KL and momentum signals,
(3) suppressing deterministic exploitation via entropy-gated penalties, and (4) constraining
search to regions closer to the reference policy. However, this analysis is entirely
mechanistic and does \emph{not} constitute empirical validation. SAFE does not address the
fundamental mismatch between the learned reward model \(r(x,y)\) and the true preference
function. At best, it may slow or redirect exploitation, but it cannot eliminate reward
hacking without improving reward model fidelity.

A rigorous assessment of SAFE’s reward-hacking behavior requires dedicated experiments with
explicitly hackable reward functions, adversarial and out-of-distribution prompts, and long
training horizons. Designing and running such evaluations are an important direction for
future work and is beyond the scope of the present study.

\end{document}